\documentclass[11pt]{article}
\usepackage{nodalida2025}
\usepackage{times}
\usepackage[table,xcdraw]{xcolor}
\definecolor{darkblue}{rgb}{0, 0, 0.5}
\usepackage[colorlinks=true, citecolor=darkblue, linkcolor=darkblue, urlcolor=darkblue]{hyperref}
\usepackage{url}
\usepackage{latexsym}
\usepackage{multicol}
\usepackage{multirow}
\usepackage{booktabs}
\usepackage{tabularx}
\usepackage{arydshln}
\usepackage{graphicx}
\usepackage[table]{xcolor}
\usepackage{comment}

\usepackage{makecell}

\aclfinalcopy % Uncomment this line for the final submission

\title{Benchmarking Abstractive Summarisation:\\ A Dataset of Human-authored Summaries of Norwegian News Articles}

\author{Samia Touileb$^1$,
  Vladislav Mikhailov$^2$, 
  Marie Kroka$^1$,
  Lilja Øvrelid$^2$,
  Erik Velldal$^2$\\
  \\
$^1$University of Bergen, $^2$University of Oslo, \\
\texttt{samia.touileb@uib.no, vladism@ifi.uio.no,}\\
\texttt{liljao@ifi.uio.no, erikve@ifi.uio.no}
  }
 
\date{}

\begin{document}
\maketitle
\begin{abstract}
We introduce a dataset of high-quality human-authored summaries of news articles in Norwegian\footnote{\url{https://github.com/SamiaTouileb/NorSumm/tree/main} and \url{https://huggingface.co/datasets/SamiaT/NorSumm/tree/main}}. The dataset is intended for benchmarking the abstractive summarisation capabilities of generative language models. Each document in the dataset is provided with three different candidate gold-standard summaries written by native Norwegian speakers, and all summaries are provided in both of the written variants of Norwegian -- Bokmål and Nynorsk. The paper describes details on the data creation effort as well as an evaluation of existing open LLMs for Norwegian on the dataset. We also provide insights from a manual human evaluation, comparing human-authored to model-generated summaries. Our results indicate that the dataset provides a challenging LLM benchmark for Norwegian summarisation capabilities. 
\end{abstract}

\section{Introduction}
One of the key practical use cases of large language models (LLMs), is to generate condensed summaries of texts. Several news publishers already include LLM-generated summaries as part of the news stories they publish.  Evaluating such generated summaries, however, remains a challenge. For Norwegian, one important reason for this is the lack of gold-standard summaries to compare to. The current paper introduces a new and open dataset of high-quality human-authored summaries of news articles in Norwegian, covering both of the official written variants; Bokmål (BM) and Nynorsk (NN). Aiming to make benchmarking as robust as possible, each document in the dataset is provided with three different candidate gold-standard summaries (for each variant, BM and NN, resulting in six summaries in total for each news article). 

The remainder of the paper is structured as follows. We first describe the creation of the human-authored summaries, including the underlying data sources, the annotator guidelines, and corpus statistics. 
We then move on to describe a first set of experiments with using pre-trained LLMs to generate summaries, and then evaluate them using our new dataset. Importantly, we here also present the methodology and framework we use, including factors like prompts and metrics. We thereafter discuss in detail the setup and results of our manual human evaluation.

\section{Related work}
\label{sec:related}

Summarisation datasets are foundational for advancing the development of techniques for automatic summarisation, as well as for benchmarking LLMs. There are various approaches developed to address diverse summarisation challenges, along with influential datasets to benchmark both extractive and abstractive methods \cite{dong2022survey,el2021automatic}. Most works on benchmark datasets have been done for English, and we here mention some of the works that focus on summarising news articles. 

The CNN/Daily Mail dataset \cite{hermann2015teaching} is one of such influential works. This dataset was created for the task of reading comprehension, but is widely used for summarisation-related tasks. The dataset consists of news articles accompanied by a set of bullet points representing (abstractive) summaries. Subsequent works have focused on creating resources for various domains, contexts, and summarisation styles. For instance, Gigaword \cite{rush-etal-2015-neural} is extracted from the Gigaword news corpus and contains sentences paired with short summaries (headlines). This is also an abstractive dataset enabling sentence-level summarisation. It has however been criticised for only including headlines instead of full summaries \cite{el2021automatic}. The extreme summarisation (XSum) dataset \cite{narayan-etal-2018-dont} was also created from news articles, sourced from BBC. Each article in this dataset is paired with a one-sentence summary representing a concise and abstractive summary. The CNN-corpus \cite{lins2019cnn} contains news articles from CNN paired with highlights and gold-standard abstractive summaries. However, the corpus is mostly used for extractive summarisation tasks \cite{el2021automatic}. Efforts have also been made for multi-document summarisation, such as Multi-News \cite{fabbri-etal-2019-multi}, which contains relatively long summaries of multi-news articles covering the same topic. 

Resources for news summarisation in Norwegian are notably scarce. Some efforts to introduce summarization datasets in Norwegian have relied on machine translation, e.g. based on the CNN/DailyMail data  \cite{liu2024nlebenchnorglmcomprehensiveempiricalanalysis}. However, failing to adequately capture nuances of the target language, as machine translation may produce non-idiomatic and non-natural-sounding language. Another concern is that, being based on English  sources, the original texts are typically not geared towards issues of primary salience to a Norwegian context (whether socially, politically, geographically, or otherwise), 
which is unfortunate if the goal is to benchmark Norwegian LLMs.

To our knowledge, no freely available, manually curated summarization dataset, created from scratch for Norwegian news data exists, making this work a valuable contribution to advancing research in this field.

\section{Human authored summaries}
\label{sec:human}

\paragraph{Data sources}

We use a subset of the news articles in the Norwegian event extraction dataset EDEN \cite{touileb-etal-2024-eden} as the data source for summarisation. EDEN contains articles in BM only, and because creating summaries based on news articles is a time and effort intensive task, we here only make use of the dev and test splits of EDEN, which respectively contain 30 and 33 news articles. EDEN was chosen due to its high-quality, as it comprises news articles from the Norwegian Dependency Treebank \cite{solberg2014norwegian, ovrelid2016universal}, and is a richly annotated dataset covering event triggers and arguments \cite{touileb-etal-2024-eden}, named entities \cite{jorgensen2019norne}, morphosyntactic annotation, and co-reference information \cite{maehlum2022narc}.

\paragraph{Annotators} 

We hired three annotators with strong academic backgrounds related to journalism, all Norwegian native speakers. The annotators were fairly compensated following an hourly contract, and were hired for a period of 6 months.  All annotators have a background in media science or journalism. The first annotator, has a bachelor's degree in media and communication science, and has worked as a freelance journalist. The second annotator has a bachelor's degree in journalism, and was finishing up a master's degree in investigative journalism while doing an internship in a leading Norwegian news broadcasting company. The third annotator, a journalism student, who also worked part-time as a journalist in a local Norwegian newspaper. 

All hired annotators have experience writing news articles, including the identification of key information that should be selected to write the article. As the task was to create natural-sounding summaries that preserved the original meaning of the news articles, we believe that these annotators can be referred to as domain experts.
In addition, as we wanted the summaries to be as natural-sounding as possible, we asked the annotators to write in their preferred variant of Norwegian. This has resulted in two annotators writing in BM, and one annotator writing in NN. 

\begin{table*}[h!]
    \centering
    \small
    \begin{tabular}{p{0.02\textwidth}|p{0.29\textwidth}p{0.29\textwidth}p{0.29\textwidth}}
    \toprule
    \multicolumn{4}{c}{\textbf{News article}} \\
    \midrule
    \multicolumn{1}{l}{} & \multicolumn{3}{p{0.9\textwidth}}{Mer frukt, men mindre norsk $|$ Forbruket økt med 20 prosent på ti år. Forbruket av frukt og grønt har økt med over 20 prosent i løpet av de siste ti årene. Men den norske produksjonen faller. Hele veksten og mer til av frukt og grønt kommer fra import. Den norske produksjonen har nemlig falt med 10 prosent siden 1998, skriver Nationen. Ifølge landbruks- og matminister Lars Peder Brekk (Sp) må de norske kjedene bli flinkere til å samarbeide med norske produsenter og bøndene må bli flinkere til å produsere det kundene vil ha.} \\
    \midrule
    \midrule
    & Summary 1 & Summary 2 & Summary 3 \\
    \midrule

    \textbf{\multirow{14}{*}{\rotatebox[origin=c]{90}{Bokmål}}} &     
    Forbruket av frukt og grønt har økt med 20 prosent på ti år. \newline Hele veksten og mer til av frukt og grønt kommer fra import da den norske produksjonen har falt med 10 prosent siden 1998, skriver Nationen. \newline De norske kjedene må bli flinkere til å samarbeide med norske produsenter og bøndene må bli flinkere til å produsere det kundene vil ha ifølge landbruks- og matminister Lars Peder Brekke (Sp). 
    & Nordmenn kjøper mer frukt og grønnsaker, samtidig som de norske bøndene produserer mindre. \newline Frukt- og grønt-forbruket har økt med over 20 prosent de 10 siste årene. \newline Den norske produksjonen har falt med 10 prosent siden 1998. \newline Import av varer dekker den økte etterspørselen i det norske markedet.
    & Forbruket av frukt og grønt har økt med 20 prosent i Norge de siste ti årene. \newline Likevel falt den norske eksporten, og veksten kommer fra stadig mer import. \newline Siden 1998 har den norske produksjonen falt med 10 prosent, opplyser Nationen. \newline Landbruksministeren sier at kjedene må bli flinkere til å samarbeide med norske produsenter, og at bøndene i større grad må produsere det kundene ønsker. \\
    
    \midrule
    
    \textbf{\multirow{14}{*}{\rotatebox[origin=c]{90}{Nynorsk}}} &     
    Forbruket av frukt og grønt har auka med 20 prosent på ti år. \newline Heile veksten og meir av frukt og grønt kjem frå import då den norske produksjonen har falle med 10 prosent sidan 1998, skriv Nationen. \newline Dei norske kjedane må bli flinkare til å samarbeide med norske produsenter og bøndene må bli flinkare til å produsera det kundene vil ha, ifølge landbruks- og matminister Lars Peder Brekke (Sp).
    & Nordmenn kjøper meir frukt og grønnsaker, samtidig som dei norske bøndene produserer mindre. \newline Frukt- og grønt-forbruket har auka med over 20 prosent dei 10 siste åra. \newline Den norske produksjonen har gått ned med 10 prosent sidan 1998. \newline Import av varer dekker den auka etterspørselen i den norske marknaden.
    & Forbruket av frukt og grønt har auka med 20 prosent i Noreg dei siste ti åra. \newline Likevel fell den norske eksporten, og veksten kjem frå meir og meir import. \newline Sidan 1998 har nemleg den norske produksjonen falle med 10 prosent, opplyser Nationen. \newline Landbruksministeren seier at kjedene må bli flinkare til å samarbeida med norske produsentar, og at bøndene må i større grad produsera kva kundane ynskjer. \\ 
         
    \bottomrule
    \end{tabular}
    \caption{Example of a news article and the summaries written by three different native speakers in either Bokmål (BM) or Nynorsk (NN), and translated into the other respective variety.}
    \label{tab:example_human_summaries}
\end{table*}

\paragraph{Guidelines}

The annotators received a detailed set of guidelines outlining the steps to follow when authoring the summaries. The guidelines were inspired by concrete prompts, shared with us, and which were used to automatically generate summaries of news articles by one of the biggest media companies in Norway. 

We asked the annotators to write summaries that reflect the main content of the news articles, but without providing strong limitations to their language use or formulations. We aimed to create summaries that are as natural-sounding as possible, and as diverse as possible. Each annotator was free to write their own summaries, without consulting or discussing details about the content of the summaries. However, we provided the annotators with an example consisting of a news article paired with its summary to discuss the format and exemplify the concrete guidelines. 

More concretely, the guidelines we provided the annotators are as follow: 
\begin{itemize}\itemsep0.01em
    \item Make a short and precise summary.
    \item The summary should be formatted as a bulleted list, with each point on a single line.
    \item The language must be clear, precise, concise, and easy to understand.
    \item Journalistic integrity must be maintained, ensure that no errors are introduced. 
    \item The summary must address the following questions: who, what, where, when, and why it is important to have knowledge of the case or event presented in the news article.
    \item The summary must be engaging and highlight key information from the article.
    \item The summary should have a maximum character count of 700, including spaces. 
\end{itemize}

We intentionally decided to keep the annotation guidelines simple to give annotators the freedom to write in a natural and authentic style. Rather than imposing strict constraints, we provided them with general and broad instructions on the importance of maintaining journalistic integrity while clearly, precisely, and concisely creating an informative summary. We believe that this flexibility allowed annotators to create more natural and engaging summaries. Our choice of enforcing summaries formatted as bullet-points was in part based on how news outlets present machine-generated summaries in the Norwegian news. But also because we planned to perform a human evaluation where human-authored summaries will be compared to machine-generated summaries. See Section \ref{sec:human_eval} for more details about this analysis.

\paragraph{Generation and evaluation}

The annotation process was carried out using a simple text editing platform, to provide the annotators a more straightforward and user-friendly interface. We had several meetings with the annotators to discuss the process and the progression of the task. However, we never aimed for aligning the content of the human-authored summaries. This was an intentional decision to create a benchmark dataset with diversity, as we believe that in the case of summarisation, there is no unique gold summary version. We wanted to create a resource that would provide three diverse summaries for each news article, in each of the written variants BM and NN.

The annotation was conducted in two rounds: (i) creating human-authored summaries, (ii) translating human-authored summaries. As previously mentioned, we gave the annotators the liberty to write in their preferred Norwegian written variant. This was to both ensure the creation of naturally-sounding summaries, but also to create a benchmark for both BM and NN. 

In the first round of annotation, our three annotators authored 63 summaries each (30 from the dev split of EDEN and 33 from the test split), following our annotation guidelines. For the second round of annotation, two of our annotators translated all summaries from BM to NN, and vice versa. Here again, the annotators translated summaries to their preferred Norwegian variant.

Since translations between the two written variants were performed by another annotator, each human-authored summary has been seen and analysed by two different annotators. We believe that this enhances the quality of the summaries, as potential ambiguity or errors could be discovered and corrected in both versions. This process again allowed us to create additional human-authored summaries for each of BM and NN. We provide more details about the resulting dataset bellow.

\paragraph{Examples} 

Table \ref{tab:example_human_summaries} shows three summaries  originally written in either Bokmål or Nynorsk, and translated into the other respective variety.

Each summary varies in both content and length, with \textit{Summary 1} being the longest and \textit{Summary 2} being the shortest (in terms of tokens). We believe that this diversity contributes to a benchmark dataset that more accurately reflects the complexities of generated summaries. Each summary presents the news article in a unique way, emphasising different important aspects of the case discussed in the news. 

The human-authored summaries exhibit differences in style and news interpretation. Some summaries are more concise, presenting only essential facts (\textit{Summary 1}), while others have a more narrative style (\textit{Summary 2} and \textit{Summary 3}) providing more contextual details. Furthermore, the summaries emphasise on varying aspects, with some focusing on key events (\textit{Summary 1} and \textit{Summary 2}), while other highlight implications or underlying causes (\textit{Summary 3}). 
We believe that this variation make our summarisation benchmark dataset more representative, and enables model evaluation on a diverse set of summaries.

\begin{table}[t!]
    \centering
    \small
    \begin{tabular}{@{}lcrrrr@{}}
    \toprule
    & Ann. & \#Summ. & \#Sent & \#Tokens & Avg.\\
    \midrule
     \multirow{ 4}{*}{\textbf{BM}}
     & A1 & 63 & 365 & 6,695 & 106.26\\
     & A2 & 63 & 280 & 6,221 & 98.74 \\
     & A3 & 63 & 312 & 6,472 & 102.73\\
     % \cmidrule(lr){2-2} \cmidrule(lr){3-3} \cmidrule(lr){4-4} \cmidrule(lr){5-5} \cmidrule(lr){6-6}
     &  & 189 & 957 &  19,042 & 102.58\\
     \midrule
     \multirow{ 4}{*}{\textbf{NN}}
     & A1 & 63 & 365 &  6,843  & 108.61 \\
     & A2 & 63 & 280 & 6,280  & 99.68 \\
     & A3 & 63 & 312 & 6,459  & 102.52 \\
    % \cmidrule(lr){2-2} \cmidrule(lr){3-3} \cmidrule(lr){4-4} \cmidrule(lr){5-5} \cmidrule(lr){6-6}
     & & 189 & 957 &  19,582 &  103.60\\
     \midrule
     \textbf{Total} & & 378 & 1,914 & 38,624 & 102,17\\
     \midrule
     \textbf{\#Doc.} & &  & 3,136 & 49,003 & 778.92 \\
     \bottomrule
    \end{tabular}
    \caption{Dataset statistics of the human-authored summaries. Left to right, the columns show language variety (Bokmål/Nynorsk),  total number of summaries,  documents, sentences, and tokens, and finally average token length of summaries. The bottom row shows the corresponding numbers for the original news articles.}
    \label{tab:counts_dataset}
\end{table}

\begin{figure*}[t]
    \centering
    \includegraphics[width=0.7\textwidth]{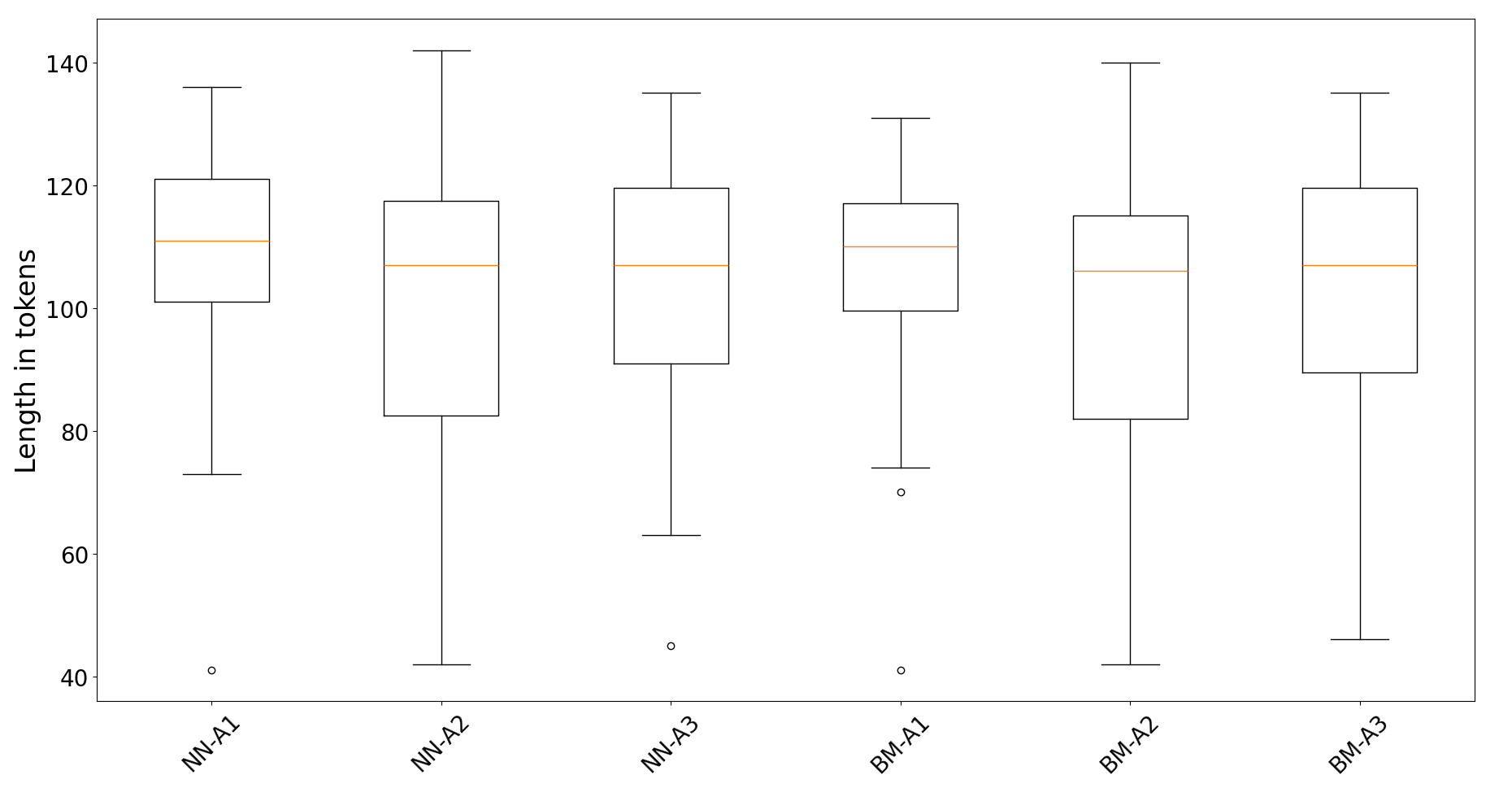}
    \caption{Box plots of summary lengths authored by three different annotators (referred to as A1, A2, and A3) in either Bokmål (BM) or Nynorsk (NN).}
    \label{fig:box:plot}
\end{figure*}

\paragraph{Dataset statistics}

As previously mentioned, our dataset uses the dev and test splits of the EDEN dataset \cite{touileb-etal-2024-eden} comprising documents written in Norwegian BM. Given the limited number of documents in each split (30 in dev and 33 in test) we present the dataset statistics as a whole, for the entire summarisation benchmark, disregarding original splits. This decision aligns with the intended usage of the dataset as a comprehensive benchmark, where treating these splits separately is not meaningful. 

Table \ref{tab:counts_dataset} shows the main statistics of our summarisation benchmark datasets, in terms of number of summaries, sentences, tokens, and average number of tokens, broken down by annotator (A1, A2, and A3) and variety (BM or NN). We also provide the total number of sentences, tokens, and average number of tokens in the original news articles for comparison. As can be seen, the total number of summaries and sentences is equal across annotators and Norwegian variety, as all annotators created summaries (and their translations) of every news article from EDEN dev and test splits. However, the number of tokens and the average number of tokens per summary varies between the language varieties and the human annotators. The first annotator (A1 in the table) has authored longer summaries than the other annotators, with annotator 2 creating the shortest ones.

These observations are clearer in the box plot in Figure \ref{fig:box:plot}, where A1, A2, and A3 refer to our three human annotators, and BM and NN are the two Norwegian varieties. The figure presents the distribution of summary (token-) lengths across the three annotators, and across the BM and NN varieties. Each annotator's summaries exhibit a range of lengths, allowing us to observe both individual tendencies and variations. The longest summary was written by annotator 1, while the shortest was written by annotator 2. The median lengths across all summaries are relatively similar, with lengths around 100-token. This we believe suggests a level of consistency in summary length, that also aligns with the guidelines given to the annotators.

There are also clear differences in term of ranges. For instance, NN-A2 has a broader range in summary lengths compared to the others, which might indicate variance in the level of details provided in the summaries. In contrast, both NN-A1 and NB-A1 display narrower ranges, implying that these summaries are more uniform in length with fewer cases of extreme variations. 

The whiskers also vary in length, with NB-A3 exhibiting particularly long whiskers. This suggest a broader range of token counts in the summaries, potentially reflecting a less standardised approach to summarisation. Outliers are observed in NN-A2, NN-A2, and NB-A1, and which indicate the presence of significantly shorter summaries than the main distribution. These outliers might represent instances of summaries that are either very condensed, lacking details or depth, or simply based on shorter original news articles.

Overall, the differences between the summaries are subtle, but still noteworthy. Summaries written by annotator 1 appear to have less variability in length, indicating greater consistence in the summarisation style. Annotator 2 seems to have a less strict and rigid way of writing summaries, which might be depending on the original length of the news article. This diversity in summary length and variability makes the datasets more natural. This suggests that models evaluated on this benchmark would need to handle varying levels of details and conciseness that necessitate the ability to meet different summarisation styles effectively.  
% Range of summaries: 
% Longest and shortest summaries from summary 1 NB =  131 41
% Longest and shortest summaries from summary 2 NB =  140 42
% Longest and shortest summaries from summary 3 NB =  135 46

\paragraph{Annotators' experience and feedback}

At the end of the annotation work, we invited annotators to reflect on their main observations and to discuss the specific aspects of the summarisation process, as well as particular news articles that they found most challenging. More concretely, we asked them to reflect on the annotation process, challenges and ambiguities in annotation, consistency in annotation, and adherence to the guidelines. With regards to the annotation process, the annotators had different strategies where for example one annotator always started by highlighting named entities, events, facts, and actions to identify the articles' main points, while another annotator read each article twice to verify accuracy and to avoid excluding details.  

Concise, bulletin-like news articles were straightforward to summarise, as their structured formats closely aligned with what they believed would constitute a good summary. 
% These typically start with a catchy introduction containing the latest and newest, and presenting further details farther down in the text.  
The annotators had a clear consensus regarding which articles were relatively straightforward to summarise and which posed greater difficulties. Sports articles and disaster-related news, injuries, or investigations were easier to summarise as they tend to contain clear and concise information. 

The annotators noted that increased complexity within certain articles directly correlated with the time required to produce high-quality summaries, highlighting the impact of article complexity on the annotation process. 
Annotators experienced that presence of subjectivity in the article was a factor indicating increased complexity. This led the annotators to make more choices, increasing the risk of making a misrepresentative summary. Examples of such ``difficult'' pieces of text are: portrait interviews, feature articles, interviews, opinion pieces, and reviews. Some articles lacked sufficient content, which required external research and made the creation of a summary more tedious. Annotators also particularly struggled with long opinion-based articles, as it was difficult for them to summarise these texts without misrepresenting opinions as facts. The longer and the more complex the article, the more difficult it was for the annotators to reduce the contents to their essence within the maximum summary size. 

All annotators reported their focus on journalistic priorities, where the aim was to convey the most relevant facts from the original news articles. While they also report a strict adherence to the guidelines, they still prioritised content accuracy over strict compliance in some cases. With regards to the translation part of the process, the annotators felt that the process was smooth and that it was easy to translate consistently.

\section{Evaluation Design}
\label{sec:exp_setup}
In the following, we illustrate the use of our summarisation dataset as an evaluation benchmark for a range of openly available Norwegian and multi-lingual LLMs.

\paragraph{Models}
We evaluate nine pretrained Transformer LLMs as our baselines: 
NorwAI-Mistral-7B\footnote{\href{https://huggingface.co/NorwAI/NorwAI-Mistral-7B}{\texttt{hf.co/NorwAI/NorwAI-Mistral-7B}}}, NORA.LLM (NorBLOOM-7B-scratch\footnote{\href{https://huggingface.co/norallm/norbloom-7b-scratch}{\texttt{hf.co/norallm/norbloom-7b-scratch}}}, NorMistral-7B-scratch\footnote{\href{https://huggingface.co/norallm/normistral-7b-scratch}{\texttt{hf.co/norallm/normistral-7b-scratch}}}, and NorMistral-7B-warm\footnote{\href{https://huggingface.co/norallm/normistral-7b-warm}{\texttt{hf.co/norallm/normistral-7b-warm}}}; \citealp{samuel2025small}), NorwAI-Llama2-7B\footnote{\href{https://huggingface.co/NorwAI/NorwAI-Llama2-7B}{\texttt{hf.co/NorwAI/NorwAI-Llama2-7B}}}, Viking-7B\footnote{\href{https://huggingface.co/LumiOpen/Viking-7B}{\texttt{hf.co/LumiOpen/Viking-7B}}}, 
Viking-13B\footnote{\href{https://huggingface.co/LumiOpen/Viking-13B}{\texttt{hf.co/LumiOpen/Viking-13B}}}, 
Mistral-7B-v.01\footnote{\href{https://huggingface.co/mistralai/Mistral-7B-v0.1}{\texttt{hf.co/mistralai/Mistral-7B-v0.1}}} \cite{jiang2023mistral7b}, 
and falcon-7b\footnote{\href{https://huggingface.co/tiiuae/falcon-7b}{\texttt{hf.co/tiiuae/falcon-7b}}} \cite{falcon40b}. All the LLMs' weights are taken from the \texttt{Transformers} library \cite{wolf-etal-2020-transformers}. 

\begin{table*}[ht!]
    \centering
    % \scriptsize
    \resizebox{\textwidth}{!}{
    \begin{tabular}{l}
    \toprule
     \multicolumn{1}{c}{\textbf{Bokmål (BM)}}  \\ \midrule
  
    1. Skriv en oppsummering av følgende artikkel med kun noen få punkter: \texttt{\{\{article\}\}}\textbackslash nOppsummering:  \\

    \vspace{-1em}\\ %\hdashline \vspace{-1em}\\
    
    2. Oppsummer følgende artikkel med noen få setninger: \texttt{\{\{article\}\}}\textbackslash nOppsummering: \\

    \vspace{-1em}\\ %\hdashline \vspace{-1em}\\
    
    % \makecell[l]{
    3. \texttt{\{\{article\}\}}\textbackslash nSkriv en kort og presis oppsummering av teksten over. Språket må være klart og lett å forstå. Sørg for å ikke introdusere feil. \\ Oppsummeringen må dekke følgende spørsmål: hvem, hva, hvor, når, og hvorfor er denne saken viktig å vite om.  Oppsummeringen må være \\ engasjerende og fremheve nøkkelinformasjon fra artikkelen. Oppsummeringen skal inneholde maksimalt 700 tegn, inkludert mellomrom.
    % }
    \\
    % \vspace{-1em}\\ \hdashline \vspace{-1em}\\
 
    4. Gi et kortfattet sammendrag av følgende tekst: \texttt{\{\{article\}\}}\textbackslash n \\
    % \vspace{-1em}\\ \hdashline \vspace{-1em}\\
    
    5. Lag en kort oppsummering som sammenfatter den følgende teksten i noen få punkter:\textbackslash n\texttt{\{\{article\}\}}\textbackslash n\textbackslash nOppsummering: \\
    % \vspace{-1em}\\ \hdashline \vspace{-1em}\\
    
    6. Heile artikkelen:\textbackslash n\texttt{\{\{article\}\}}\textbackslash n\textbackslash nHovudpunkt: \\

    \midrule 
    \multicolumn{1}{c}{\textbf{Nynorsk (NN)}} \\
    \midrule 
    
    1. Skriv ei oppsummering av følgande artikkel med berre nokre få punkt: \texttt{\{\{article\}\}}\textbackslash nOppsummering: \\
    % \vspace{-1em}\\ \hdashline \vspace{-1em}\\
    
    2. Oppsummer følgande artikkel med nokre få setningar: \texttt{\{\{article\}\}}\textbackslash nOppsummering: \\
    % \vspace{-1em}\\ \hdashline \vspace{-1em}\\
    
    % \makecell[l]{
        3. \texttt{\{\{article\}\}}\textbackslash nSkriv ein kort og presis oppsummering av teksten over. Språket må vere klart og lett å forstå. Sørg for å ikkje introdusere feil. \\ Oppsummeringa må dekkje følgande spørsmål: kven, kva, kor, når, og kvifor er denne saka viktig å vite om.  Oppsummeringa må vere \\ engasjerande og framheve nøkkelinformasjon frå artikkelen. Oppsummeringa skal innehalde maksimalt 700 tegn, inkludert mellomrom.
    %} 
    \\
    % \vspace{-1em}\\ \hdashline \vspace{-1em}\\
    
    4. Gje eit kortfatta samandrag av følgande tekst: \texttt{\{\{article\}\}}\textbackslash n \\
    
    % \vspace{-1em}\\ \hdashline \vspace{-1em}\\
    5. Lag ein kort oppsummering som samanfattar den følgande teksten i nokre få punkt:\textbackslash n\texttt{\{\{article\}\}}\textbackslash n\textbackslash nOppsummering: \\
    % \vspace{-1em}\\ \hdashline \vspace{-1em}\\
    
    6. Hele artikkelen:\textbackslash n\texttt{\{\{article\}\}}\textbackslash n\textbackslash nHovedpunkter:
    
    \\

    \midrule 
    \multicolumn{1}{c}{\textbf{English translation}} \\
    \midrule 

     1. Write a summary of the following article in just a few points: \texttt{\{\{article\}\}}\textbackslash nSummary:  \\

    \vspace{-1em}\\ %\hdashline \vspace{-1em}\\
    
    2. Summarise the following article in a few sentences: \texttt{\{\{article\}\}}\textbackslash nSummary: \\

    \vspace{-1em}\\ %\hdashline \vspace{-1em}\\
    
    % \makecell[l]{
    3. \texttt{\{\{article\}\}}\textbackslash nWrite a short and precise summary of the text above. The language must be clear and easy to understand. Ensure not to introduce errors. \\ The summary must cover the following questions: who, what, where, when, and why this matter is important to know about. The summary must be \\ engaging and highlight key information from the article. The summary should contain a maximum of 700 characters, including spaces.
    % }
    \\
    % \vspace{-1em}\\ \hdashline \vspace{-1em}\\
 
    4. Provide a concise summary of the following text: \texttt{\{\{article\}\}}\textbackslash n \\
    % \vspace{-1em}\\ \hdashline \vspace{-1em}\\
    
    5. Create a short summary that encapsulates the following text in a few points:\textbackslash n\texttt{\{\{article\}\}}\textbackslash n\textbackslash nSummary: \\
    % \vspace{-1em}\\ \hdashline \vspace{-1em}\\
    
    6. The entire article:\textbackslash n\texttt{\{\{article\}\}}\textbackslash n\textbackslash nMain point: \\
    \bottomrule
    \end{tabular}
    }
    \caption{Six prompts in both BM and NN used in our zero-shot evaluation experiments (\S\ref{sec:results}).}
    \label{tab:prompts}
\end{table*}

% \textbackslash n

\paragraph{Setup} We conduct a zero-shot evaluation of the previously mentioned LLMs using \texttt{noreval}\footnote{\href{https://github.com/ltgoslo/noreval}{\texttt{github.com/ltgoslo/noreval}}}, an open-source framework for evaluating Norwegian generative LLMs. We integrate our dataset into \texttt{noreval} together with 12 diverse prompts written by Norwegian native speakers, who are authors of this paper. \autoref{tab:prompts} illustrates the prompts -- 6 prompts per language variety. As can be seen, we use a variety of prompting styles to generate summaries, varying both the placement of the source article, as well as the verbosity and precise wording of the instruction. The LLMs' summaries are generated via the greedy search decoding method.

\paragraph{Performance Metrics}
We measure the performance using standard summarisation evaluation metrics: ROUGE-L~\cite{lin-2004-rouge} and BERTScore \cite{zhang2019bertscore}. Our result aggregation procedure accounts for prompt sensitivity \cite{voronov2024mind,lu-etal-2024-prompts} and includes two steps: (i) for each prompt, we compute the maximum performance scores between the LLM's output and each of three human-written references (our human-authored summaries); (ii) we then maximize the BERTScore across all prompts and average the resulting ROUGE-L and BERTScore values over all BM/NN examples.

\section{Evaluation Results}
\label{sec:results}
\begin{table*}[ht!]
    \centering
    \scriptsize
    \resizebox{\textwidth}{!}{ % 
    \begin{tabular}{lcccccc}
\toprule
\multirow{2}{*}{\textbf{Model}} &  \multicolumn{2}{c}{\textbf{BM}}  &  \multicolumn{2}{c}{\textbf{NN}} &  \multicolumn{2}{c}{\textbf{Overall}} \\ \cmidrule{2-7}
   &   \textbf{ROUGE-L}      &      \textbf{BERTScore}      &     \textbf{ROUGE-L}    &     \textbf{BERTScore} & \textbf{ROUGE-L}    &     \textbf{BERTScore}      \\
\midrule
NorwAI-Mistral-7B          &   12.14 &      50.06 &   10.62 &      50.78 & 11.38 & 50.42 \\
%NorwAI-Mistral-7B-pretrain &   10.62 &      49.60 &   11.71 &      50.84 \\
NorwAI-Llama2-7B           &   13.58 &      54.44 &   12.24 &      54.04 & 12.91 & 54.24 \\
norbloom-7b-scratch        &   20.00 &      52.40 &   13.29 &      49.16 & 16.6 & 50.78 \\
normistral-7b-scratch      &   25.32 &      58.25 &   15.28 &      48.32 & 20.3 & 53.28 \\
normistral-7b-warm         &   17.38 &      49.86 &    9.93 &      41.86 & 13.6 & 45.86 \\
Viking-7B                  &   \underline{30.56} &      \underline{69.65} &   \underline{25.82} &      \textbf{70.34} & \underline{28.19} &  \underline{70.0} \\
Viking-13B                 &   \textbf{33.76} &      \textbf{70.90} &  \textbf{ 27.38} &      \underline{69.96} & \textbf{30.57} & \textbf{70.4} \\ 
    
    \vspace{-1em}\\ \cdashline{1-7} \vspace{-1em}\\
Mistral-7B-v0.1            &    9.60 &      52.36 &    8.70 &      47.28  & 9.15 & 49.82 \\
falcon-7b                  &   10.61 &      44.40 &    9.80 &      44.06 & 10.2 & 44.23 \\
\bottomrule
\end{tabular}}
    \caption{Zero-shot evaluation results on concatenated development and test sets by BM and NN. The best score is in bold, second-best is underlined. The LMs with more limited abilities in Norwegian are separated by a dashed line.}
    \label{tab:overall}
\end{table*}

\autoref{tab:overall} presents the zero-shot evaluation results on concatenated development and test sets. In addition to this evaluation, we conducted a human-based evaluation (see \S\ref{sec:human_eval}) to analyse the LLMs' behaviour in more detail given the limitations of the automatic performance metrics \cite{gehrmann2023repairing,colombo-etal-2023-glass}. 

\paragraph{Overall Results} We find that all LLMs achieve acceptable performance on both BM and NN. Viking-7B and Viking-13B perform the best, reaching the ROUGE-L of up to 33.76 and BERTScore of up to 70.34. The larger version is insignificantly better than the smaller one. We also observe that Norwegian monolingual LLMs (NorwAI-Mistral-7B, NorwAI-Mistral-7B-pretrain, and NorMistral-7B-warm) can perform on par with LLMs with more limited abilities in Norwegian (Mistral-7B-v0.1 and Falcon-7b). 

\paragraph{Comparison of BM \& NN} Comparing the results between BM and NN, we find that most LLMs performs better on BM in terms of ROUGE-L (e.g., the $\delta$-score ranges from 1 to 10 for NorwAI-Mistral-7B-pretrain and NorBLOOM-7B-scratch, respectively). However, the BERTScore difference is less pronounced.

The relatively low performance scores suggest that our summarisation dataset presents a challenging benchmark. One could argue that using more advanced, proprietary LLMs, which have demonstrated higher effectiveness in summarisation tasks, could yield better results than the models we have evaluated here. However, we chose to rely exclusively on open-source models with Norwegian language support to ensure accessibility and reproducibility for future research.

\section{Human evaluation}
\label{sec:human_eval}

\begin{figure}[t]
    \centering
    \includegraphics[width=\linewidth]{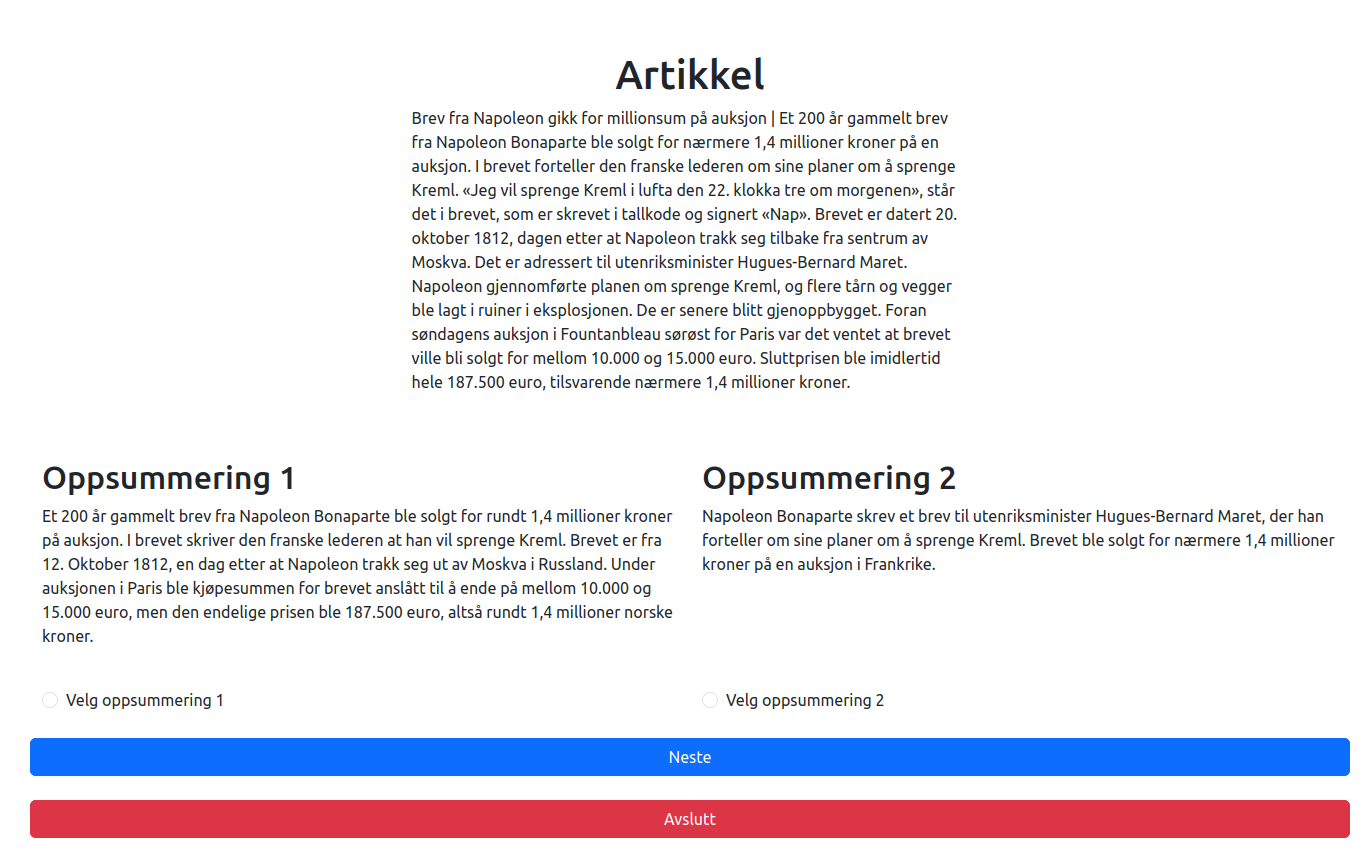}
    \caption{Screenshot of the interface used during human evaluation. We present a news article on top, and two suggestions for summaries. The goal for the evaluator is to choose the summary they prefer based on simple criteria (see \S\ref{sec:human_eval}).}
    \label{fig:example_annotation_interface}
\end{figure}

In addition to model and metric-based evaluations, we conducted a manual evaluation. For this purpose, a research assistant was hired to develop an interface where evaluators were shown a news article, followed by two summaries beneath it. An example of this simple interface is shown in Figure \ref{fig:example_annotation_interface}. The volunteer evaluators were asked to choose their preferred summary from a selection of two summaries: one human-authored and one generated by a model. However, the evaluators were not aware of the provenance of each summary. 

To ensure that evaluators rank summaries consistently, we provided them with a set of very simple criteria inspired by evaluations presented in \cite{fabbri-etal-2021-summeval}:
\begin{itemize}\itemsep0.2em
    \item \textbf{Relevance}: Selection of essential content from the original news article.
    \item \textbf{Consistency}: Alignment between the summary and the source article, ensuring that the summary contains only factual statements that can be directly inferred from the source.
    \item \textbf{Fluency}: Quality of individual sentences, with particular attention to grammatical correctness to ensure readability.
\end{itemize}
    
We also asked the evaluators to prioritise these criteria in the following order: relevance $>$ consistency $>$ fluency, with relevance being the most important and fluency the least. This approach was designed to assess the quality of the summaries based on the primary functions of summarisation: accurately and concisely conveying essential content. The prioritisation we chose reflects a deliberate emphasis on accuracy and factuality over style.

The link to this evaluation interface was shared with volunteer colleagues, resulting in a total of 146 responses. In 138 cases, evaluators preferred the human-authored summaries, while only 8 responses favoured a machine-generated summary. These preferred machine-generated summaries were produced by the three models Viking-13B (4 of the preferred summaries), NorBLOOM-7b-scratch (2 of the preferred summaries), and NorMistral-7b-warm (2 of the preferred summaries), using prompt nr. 1 (BM) and prompt nr. 2 (BM) in Table \ref{tab:prompts}.

Similarly to the results in Table \ref{tab:overall}, the best model metric-wise, Viking-13B, seem to also be the model most favoured by human evaluators. Although this preference remains limited compared to the preference of human-authored summaries, it provides an indication of the quality of summaries generated by this model compared to the others. 

Several issues were identified during the human evaluation of summaries. These were primarily related to those generated by the models. We give a summary of the types of errors that commonly appeared in what follows. 

\paragraph{Issues related to relevance} the generated summaries often reproduce the initial part of the original article, not including important information presented later, and sometimes even cutting off mid-sentence. Some summaries were direct copy-paste of the original article, or were too lengthy, and occasionally repeating (parts of) the prompts (e.g. ``Skriv en oppsummering av følgende artikkel med kun noen få punkter: Tilbake til hverdagen $|$ Helse. Vandrehall [\ldots]'', \textit{eng}: \textit{Write a summary of the following article in just a few points: Back to Everyday Life $|$ Health. Walking hall [\ldots]}). Some other summaries were too short, providing incomplete contexts or unnatural-sounding sentences.  

\paragraph{Issues related to consistency} generally, evaluators reported that the summaries were consistent with the source material. However, some summaries did exhibit repetitions of phrases. Minor but significant alterations in the texts, like adding or omitting words, were also observed. In some instances, the model-generated summaries invented quotes (e.g. a citation in the summary that did not occur in the original news text ``- Jeg er veldig glad for at jeg har fått et nytt hjerte, sier Per Arne Olsen til Tønsbergs Blad.'' (\textit{eng}: \textit{`- I am very happy that I have received a new heart, says Per Arne Olsen to Tønsbergs Blad.''}). However, a simple internet search led us to finding a similar quote in another news article which seemingly the model had access to during training), or confused entities (e.g., mixing between Bill and Hillary Clinton when mentioned jointly in a news article).

\paragraph{Issues related to fluency} similarly to what we already have mentioned, despite fluency being largely maintained, certain summaries repeated identical or similar sentences continuously (more than 10 times). Additionally, in some cases we observed missing function words (e.g. the function word ``av'' (\textit{eng}: \textit{by}) in the sentence ``Malis statsminister Cheick Modibo Diarra har gått av etter å ha blitt pågrepet soldater'' (\textit{eng}: \textit{Mali's Prime Minister Cheick Modibo Diarra has resigned after being arrested soldiers}) not being included in the same sentence in the generated summary.)

\section{Conclusion and Outlook}
This paper introduces a novel dataset of human-authored summaries of Norwegian news articles for benchmarking abstractive summarisation. Our dataset is of high quality and provides for each news article a set of diverse summaries written in both Norwegian varieties Bokmål and Nynorsk. Through comprehensive evaluations using human evaluators and generative models, we have demonstrated the robustness and complexity of this benchmark.

As this is the first freely available human-authored Norwegian summarisation datasets, we believe that the impact it will have on benchmarking current and future LLMs is considerable. Looking ahead, we see several avenues for developing models that leverage the particularities of this dataset to build more robust summarisation techniques. This dataset allows us to compare the output of generative models to a distinct set of human-authored summaries, which will allow us to generate more naturally-sounding summaries.

% This benchmark dataset will be made freely available upon acceptance of the paper. 

\section*{Acknowledgments}
We would like to thank our annotators Marie I. Kroka, Frida Måseidvåg, and Lidvard Sandven for their great work on producing the human summaries and their translated counterparts.  
This work was supported by industry partners and the Research Council of Norway with funding to MediaFutures: Research Centre for Responsible Media Technology and Innovation, through the centers for Research-based Innovation scheme, project number 309339.

\bibliographystyle{acl_natbib}
\bibliography{anthology_0,anthology_1,nodalida2025}

\end{document}